\documentclass[conference]{IEEEtran}
\usepackage{cite}

\ifCLASSINFOpdf
\usepackage[pdftex]{graphicx}
\DeclareGraphicsExtensions{.pdf,.jpeg,.png}
\usepackage{amsmath}
\usepackage{array}

\ifCLASSOPTIONcompsoc
\usepackage[caption=false,font=normalsize,labelfont=sf,textfont=sf]{subfig}
\else
\usepackage[caption=false,font=footnotesize]{subfig}
\fi
\usepackage{url}

\usepackage{booktabs}

\begin{document}
%
\title{Extracting Lungs from CT Images using Fully Convolutional Networks}

\author{\IEEEauthorblockN{Jeovane H. Alves, Pedro M. Moreira Neto, Lucas F. Oliveira}
	\IEEEauthorblockA{Laboratory of Vision, Robotics and Imaging (VRI)\\
		Department of Informatics (DInf)\\
		Federal University of Parana\\
		Curitiba, PR - Brazil\\
		Email: \{jhalves, pmmn11, lferrari\}@inf.ufpr.br}}


%


\maketitle

\begin{abstract}	
	Analysis of cancer and other pathological diseases, like the interstitial lung diseases (ILDs), is usually possible through Computed Tomography (CT) scans. To aid this, a preprocessing step of segmentation is performed to reduce the area to be analyzed, segmenting the lungs and removing unimportant regions.
	Generally, complex methods are developed to extract the lung region, also using hand-made feature extractors to enhance segmentation. 
	With the popularity of deep learning techniques and its automated feature learning, we propose a lung segmentation approach using fully convolutional networks (FCNs) combined with fully connected conditional random fields (CRF), employed in many state-of-the-art segmentation works. 
	Aiming to develop a generalized approach, the publicly available datasets from University Hospitals of Geneva (HUG) and VESSEL12 challenge were studied, including many healthy and pathological CT scans for evaluation.
	Experiments using the dataset individually, its trained model on the other dataset and a combination of both datasets were employed. Dice scores of $98.67\%\pm0.94\%$ for the HUG-ILD dataset and $99.19\%\pm0.37\%$ for the VESSEL12 dataset were achieved, outperforming works in the former and obtaining similar state-of-the-art results in the latter dataset, showing the capability in using deep learning approaches.
\end{abstract}


%
\IEEEpeerreviewmaketitle

\section{Introduction}	
Cancer is one of the leading causes of death worldwide, with estimates of cases and mortality increasing over the years in less developed countries then more developed ones \cite{Torre2015}. Among different types, the lung cancer is the leading cause of death, except less developed countries' females. Studies in the US presented a declining of lung cancer deaths in last years (mainly in males), as a result of smoking reduction and enhancement in early diagnosis and treatment \cite{siegel2018cancer}. Nevertheless, further research is needed to decrease and control cancer cases, also for other health issues, e.g. interstitial lung diseases (ILDs).

Diagnosis of health issues is possible using data of different medical scan technologies. They may present distinct information about the body (e.g. anatomical and/or functional). A wide-used medical exam is the X-ray Computed Tomography (CT), also commonly referred as Computed Tomography. A CT scan can present a 3D structure from some part of or from the entire body, commonly used for detection of body anomalies like cancer. Thoracic area is widely scanned using CT scans to analyze the lung region, for detection of nodules (benign or malignant ones) and irregularities in the parenchyma \cite{el2013computer}.

Experts analyze each CT slice (a 2D image of some depth from the 3D scan) to evaluate the patient (e.g. to found out malignant nodules). This evaluation can be exhaustive depending on the volume of scans analyzed and on its depth/complexity, which can reduce diagnostic reliability. Semi- and automatic approaches using computational techniques may aid experts, as a second opinion, and minimize the region of interest, thus reducing the number of slices evaluated. To increase the success not only of these approaches, but also to improve the evaluation of experts, a preprocessing step of segmentation is usually performed. In lung-related diseases, this step aims the extraction of the lung region only, removing unimportant and false-positive areas \cite{van2013automated}.

\subsection{Related Work}
\label{sec:related_work}

Diverse works aimed to segment the lung region using different and combined techniques, that goes through region growing \cite{weinheimer2011automatic}, border analysis \cite{rebouccas2017novel}, shape and probabilistic models \cite{sun2012automated,nakagomi2013multi,dai2015novel}, and recently deep learning approaches \cite{Harrison2017}.
A combination of histogram and morphological operations to extract the airspace, remove the bronchi and then segment the left and right lungs was proposed in \cite{lassen2011lung}, which have obtained a mean overlap score (between lung mask and segmentation) of $97.3\%$ in the LOLA11 dataset\footnote{Challenge can be accessed online at \url{https://lola11.grand-challenge.org/}}.
Another lung segmentation work \cite{weinheimer2011automatic} experimented on this dataset, with a $97\%$ mean overlap score. For experiments, they used a in-house software named \textit{yacta} ('yet another CT analyzer'), which implements a region growing based method, processing each lung side.
A novel robust active shape model (RASM) matching combined with a optimal surface finding method was proposed as a two-step approach for left and right lung segmentation \cite{sun2012automated}. Experiments with a private dataset containing 30 CT scans (healthy and diseased ones) achieved a Dice similarity coefficient (DSC) of $97.5\% \pm 0.6\%$.
Another approach is the one proposed by Nakagomi et al. \cite{nakagomi2013multi}. They presented a novel shape graph cut based lung segmentation, which the best results were achieved by a multi-shape graph cut approach with Jaccard index, also known as Intersection over Union (IoU), of $97.68\% \pm 1.05\%$. Experiments were performed with a synthetic image and 97 CT scans.

A wide evaluation on more than 400 CT scans with interstitial lung diseases and/or nodules was performed in a two-stage method based on fuzzy connectedness (FC) segmentation, detection of the rib-cage and texture features \cite{mansoor2014generic}. This approach consists in obtaining a initial segmentation of the lung parenchyma through thresholding, seed generation for the left and right lungs, then the FC segmentation. In the second stage, a random forest classification combined with space information (e.g. rib cage) was used to identify lung pathologies, hence improving the initial segmentation. Average Dice coefficient of $95.95\% \pm 0.34\%$ and $96.27\% \pm 10.58\%$ for each observer were obtained in the datasets studied (excluding the LOLA11 challenge dataset). A mean overlap up to $96.8\%$ was achieved in the LOLA11 dataset.
Another fuzzy-based segmentation is the one proposed by Zhou et al.\cite{zhou2014automated}, which aimed to extract lungs with juxtapleural nodules. After preprocessing and thorax extraction, lung segmentation is performed using a fuzzy-c-means clustering, then refinement with iterative weighted averaging and adaptive curvature threshold. A mean IoU of $95.81\%\pm0.89\%$ was achieved, also stated that every juxtapleural nodules was included in the segmentation.
Another graph cut segmentation approach is the one developed in \cite{dai2015novel}, which uses Gaussian mixture models (GMMs) and expectation maximization (EM). After a noise-removal Gaussian smoothing, a graph representing the image is constructed, then the min-cut/max-flow algorithm is applied to segment it into foreground and background. For this segmentation, seed points for both regions are required. Thus, the authors applied the GMMs with aiding of the EM-MAP (Maximum a Posteriori) algorithm for parameter estimation (although K-means clustering is used for parameter initialization). This approach had an average Dice of $98.74\%\pm0.70\%$.

Although Shen et al. proposed a bidirectional chain-code lung segmentation in \cite{shen2015automated}, its focus and evaluation were based on juxtapleural nodules found in the Lung Imaging Database Consortium (LIDC) dataset. Inflection points are detected both horizontally and vertically, based on the initial segmentation. Three features from these points were extracted and used as inputs for the SVM classifier. Then, lung border is reconstructed with the positive-labeled points.
Evaluation in 233 CT scans reported a inclusion rate of $92.6\%$ (373 out of 406 juxtapleural nodules), with averages of under- and over-segmentation of $2.4\%$ and $0.3\%$ respectively.
Also, a control feedback system based segmentation using morphology and texture was proposed in \cite{noor2015automatic} for analysis of 96 patients, which 81 had presence of ILDs. This approach obtained a Dice coefficient of $98.21\%\pm1.35\%$ and $98.58\%\pm1.28\%$ for left and right lungs respectively, with an average of $95.40\%$ for both lungs.
A proposed combination of incremental and constrained non-negative matrix factorization (NMF), named incremental constrained NMF (ICNMF), was experimented with pathological-simulated lung phantoms and also in real CT scans (a dataset with 17 cases and another from the LOLA11 challenge)\cite{hosseini20163}. Having the Dice scores of $96\%$ and $96.5\%$ (LOLA11 dataset), the technique had shown its robustness although, as stated by the authors, exclusion of nine dense pathological scans resulted in an increasing of its score to $98.6\%$, which would suggest a lower performance in cases of this pattern.

Soliman et al. \cite{soliman2017accurate} segmented the lungs through adaptive shape model, Linear Combination of Discrete Gaussian (LCDG) and Markov-Gibbs Random Fields (MGRF).
Several versions of an original CT scan are generated by Gaussian scale space filtering then, together with the original image, used as inputs for the proposed method. Each one of these scans are subjected to the generation of first (LCDG) and second-order models (spatially uniform MGRF), then combined with shape priors. At last, a majority rule merging results into the final segmentation.
Attaining Dice coefficients of $98.4\%\pm1.0\%$ and $99.0\%\pm0.5\%$, for the UoLDB (in-house) and VESSEL12 datasets respectively, and an overlap of $98.0\%\pm7.5\%$ for the LOLA11 dataset, being ranked first in this competition.
Following the popularity of deep learning models in recent years, also in segmentation with fully convolutional networks (FCN), Harrison et al. \cite{Harrison2017} proposed the P-HNN (progressive holistically-nested networks) model, a FCN-based approach which uses a progressive multi-path scheme to refine its results, aiming pathological lung segmentation. 
Experiments were conducted in three datasets: from LTRC (Lung Tissue Research Consortium), HUG-ILD and an infection-based dataset from NIH (National Institutes of Health). Respectively, Dice scores of $98.7\%\pm0.5\%$, $97.9\%\pm1.0\%$ and $96.9\%\pm3.4\%$ were attained. Table \ref{tab:summary_state_of_the_art} summarizes the works reported above.

\begin{table}[htb]
	\centering
	\caption{Summary of related works in different datasets}
	\label{tab:summary_state_of_the_art}
	\begin{tabular}{@{}lllp{5cm}@{}}
		\toprule
		Work                           & Year & Dataset  & Results                                        \\ \midrule
		\cite{lassen2011lung}          & 2011 & LOLA11   & $97.3\%$ overlap                                    \\ \midrule
		\cite{weinheimer2011automatic} & 2011 & LOLA11   & $97\%$ overlap                                         \\ \midrule
		\cite{sun2012automated}        & 2012 & Private  & DSC of $97.5\% \pm 0.6\%$                      \\ \midrule
		\cite{nakagomi2013multi}       & 2013 & Private  & IoU of $97.68\% \pm 1.05\%$                    \\ \midrule
		\cite{mansoor2014generic}      & 2014 & Various & $95.95\% \pm 0.34\%$ and $96.27\% \pm 10.58\%$ from 1st and 2nd observers \\
		&      & LOLA11   & $96.8\%$ overlap            \\ \midrule
		\cite{zhou2014automated}       & 2014 & Private & IoU of $95.81\%\pm0.89\%$                      \\ \midrule
		\cite{dai2015novel}            & 2015 & Private & DSC of $98.74\%\pm0.70\%$                      \\ \midrule
		\cite{shen2015automated}       & 2015 & LIDC     & Inclusion rate of $92.6\%$                     \\ \midrule
		\cite{noor2015automatic}       & 2015 & Private         & DSC of $98.21\%\pm1.35\%$ and $98.58\%\pm1.28\%$ for left and right lungs. Average of $95.40\%$ \\ \midrule
		\cite{hosseini20163}           & 2016 & Private  & DSC of $96\%$                                  \\
		&      & LOLA11   & DSC of $96.5\%$                               \\ \midrule
		\cite{soliman2017accurate}     & 2017 & UoLDB    & DSC of $98.4\%\pm1.0\%$                        \\
		&      & VESSEL12 & DSC of $99.0\%\pm0.5\%$                        \\
		&      & LOLA11   & DSC of $98.0\%\pm7.5\%$                        \\ \midrule
		\cite{Harrison2017}            & 2017 & LTRC     & DSC of $98.7\%\pm0.5\%$                        \\
		&      & HUG-ILD  & DSC of $97.9\%\pm1.0\%$                        \\
		&      & NIH      & DSC of $96.9\%\pm3.4\%$                        \\ \bottomrule
	\end{tabular}
\end{table}

Most works employed specific techniques which may not generalize to deviations from the studied samples, mainly without the presence of ILD cases. Although some works applied techniques in pathological lungs, further work is necessary to improve segmentation, since working on dense pathologies is still a challenge. Besides, popularity and promising results by deep learning approaches in many areas would incentive additional application in medical research. Hence, we propose a fully convolutional network based lung segmentation, which aims not only to segment healthy lungs, but also adapts to dense pathologies, correctly segmenting the lung region with application of more general and straightforward methods.

In the next sections, we present the methodology of our work and the results obtained. Section \ref{sec:data_sets} briefly describes the datasets studied. Our lung segmentation approach is presented in Section \ref{sec:methods}, then details about experiments and the results achieved by them in Section \ref{sec:experiments}. Finally, Section \ref{sec:conclusion} concludes our study.

\section{Datasets}
\label{sec:data_sets}

In this section, a briefly description of the datasets used in our experiments is presented.

\subsection{HUG-ILD}

\begin{figure}[h]
	\centering
	\subfloat{%
		\includegraphics[width=0.48\linewidth]{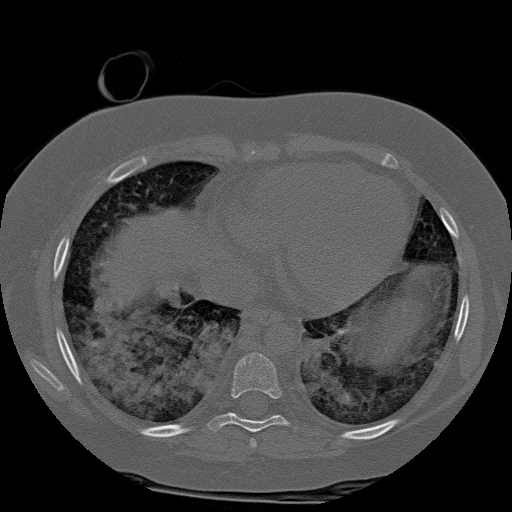}}
	\label{example_hug_ild_a}\hfill
	\subfloat{%
		\includegraphics[width=0.48\linewidth]{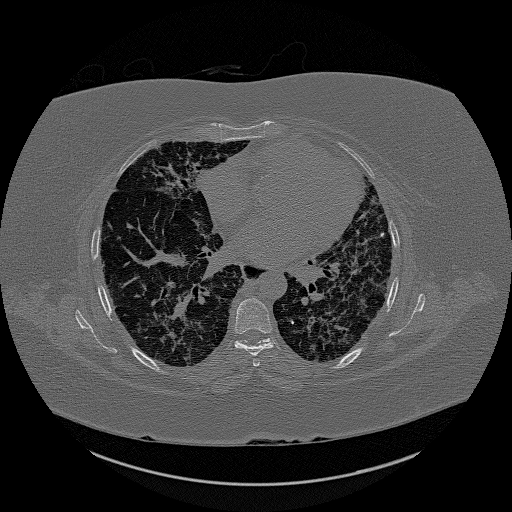}}
	\label{example_hug_ild_b}
	\caption{Examples of pathological images in the HUG-ILD dataset.}
	\label{fig:example_hug_ild} 
\end{figure}

Cases of interstitial lung diseases (ILDs) were publicly available in a dataset built at the University Hospitals of Geneva (HUG) \cite{depeursinge2012building}, motivating further studies about ILDs (in this paper, we will be referring it as HUG-ILD dataset). Compromised of 128 patients and 108 annotated CT series, the HUG-ILD dataset contains diagnosis from 13 ILDs, providing more variability. One thing to note is, despite the high number of annotated CT series, they have small quantity of slices (less than 100). Some examples are shown in Figure \ref{fig:example_hug_ild}. Also, one case had half of its ground-truths blank, so these slices were removed from the experiments.

\subsection{VESSEL12}

\begin{figure}[hbt]
	\centering
	\subfloat{%
		\includegraphics[width=0.48\linewidth]{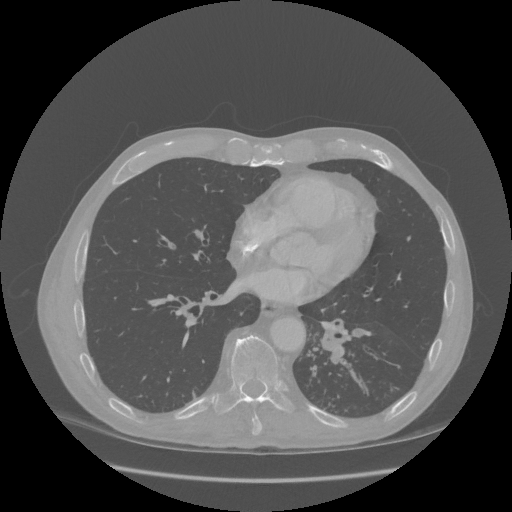}}
	\label{fig:example_vessel12_a}\hfill
	\subfloat{%
		\includegraphics[width=0.48\linewidth]{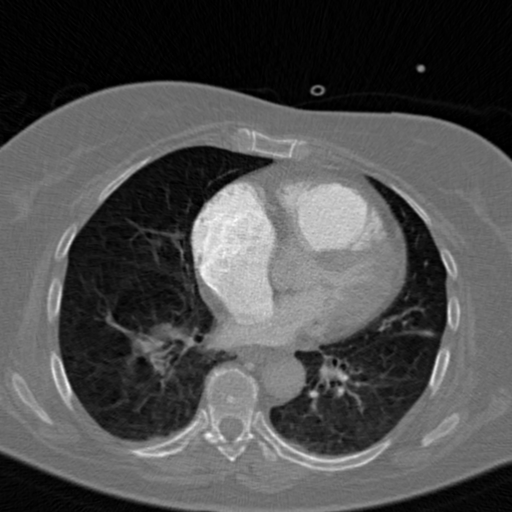}}
	\label{fig:example_vessel12_b}
	\caption{Some slices from VESSEL12 dataset.}
	\label{fig:example_vessel12} 
\end{figure}

The VESsel SEgmentation in the Lung 2012 (VESSEL12) challenge \cite{rudyanto2014comparing}\footnote{Challenge can be accessed online at \url{https://vessel12.grand-challenge.org/home/.}} was proposed for the evaluation of both semi- and automatic methods for lungs' blood vessel segmentation in CT scans. A total of 20 scans were available for testing (plus three as examples). These methods would perform vessel segmentation and then their resulted annotations submitted to an online platform for validation. To aid vessel segmentation, annotated lung masks were included. Scans have an average of 432 slices (examples in Figure \ref{fig:example_vessel12}).

\section{Lung Segmentation with FCN and CRF}
\label{sec:methods}

\begin{figure}[hbt]
	\centering
	\includegraphics[width=1\linewidth]{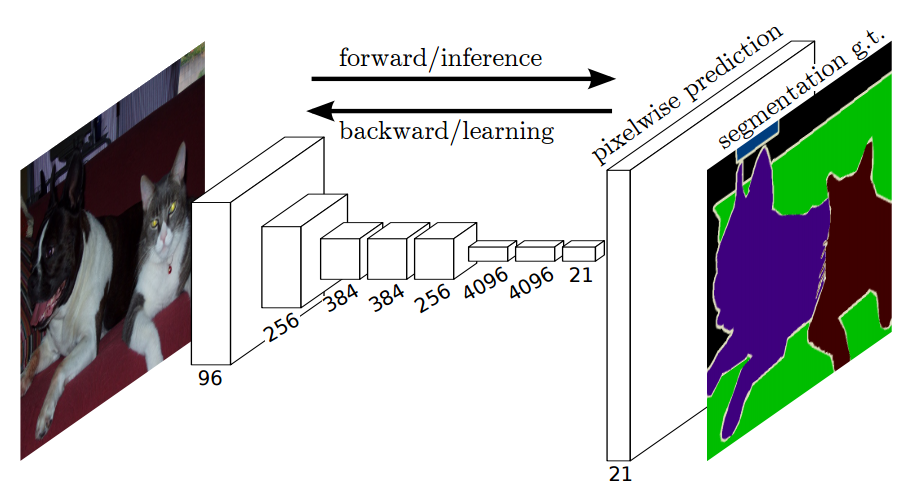}
	\caption{Semantic segmentation using convolutional networks. Extracted from \cite{shelhamer2017fully}.}
	\label{fig:fcn}
\end{figure}

In the last years, usage of deep learning models is increasing and evolving to solve computer vision problems. These problems include object detection, biometric recognition and semantic segmentation. Long et al. \cite{long2015fully} popularized the usage of convolutional networks for semantic segmentation (e.g. Figure \ref{fig:fcn}) with their work, presenting a fully convolutional network (FCN) with a skip architecture. FCN replaces the usage of fully connected layers with fully convolutional ones. Many latter segmentation works were based on this technique. Other approaches like deconvolution, unpooling, dilated convolution and multi-path refinement were also implemented in these models to enhance segmentation \cite{shelhamer2017fully}.

\begin{figure}[hbt]
	\centering
	\includegraphics[width=1\linewidth]{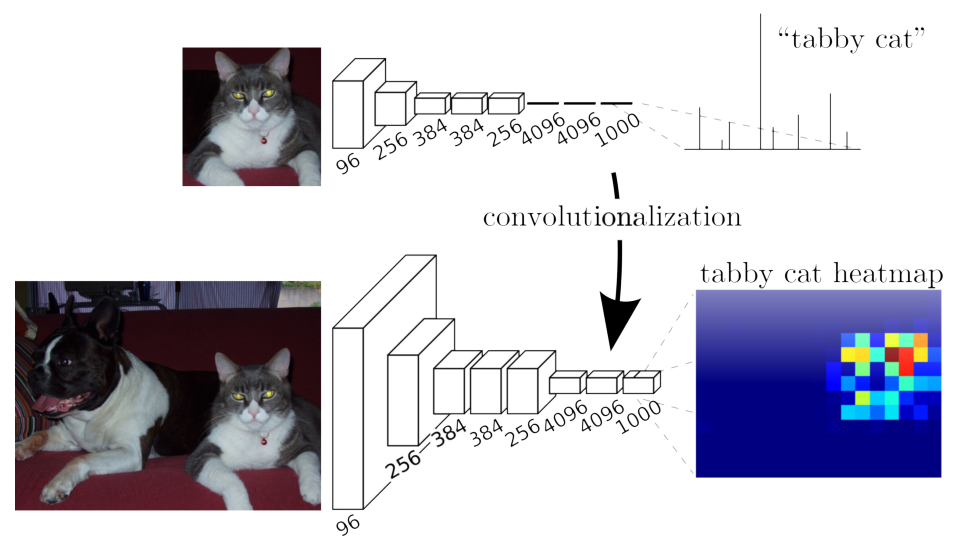}
	\caption{Example of a network with fully connected layers being transformed into a FCN. Extracted from \cite{shelhamer2017fully}.}
	\label{fig:convolutionalization}
\end{figure}

To better understand how FCNs work and their qualities, we first briefly explain what are convolutional networks. Also know as convnets, they are networks composed mainly by convolution layers, besides the presence of pooling and fully connected layers. Convnets "learn" the patterns of local regions in their convolution layers, applying convolutions in the image with a specific-size kernel, which its weights are tuned by feed-forward computation and back propagation. Also, pooling layers are employed to reduce the size of convolutional layer's output, increasing the area of effect of convolutions in latter layers, thus obtaining more global features.
FCNs are networks composed of convolution layers with the absence of fully connected ones. One can transform a convnet into a FCN by converting its fully connected layers into fully convolutional layers (e.g. shown in Figure \ref{fig:convolutionalization}).

\begin{figure}[hbt]
	\centering
	\includegraphics[width=1\linewidth]{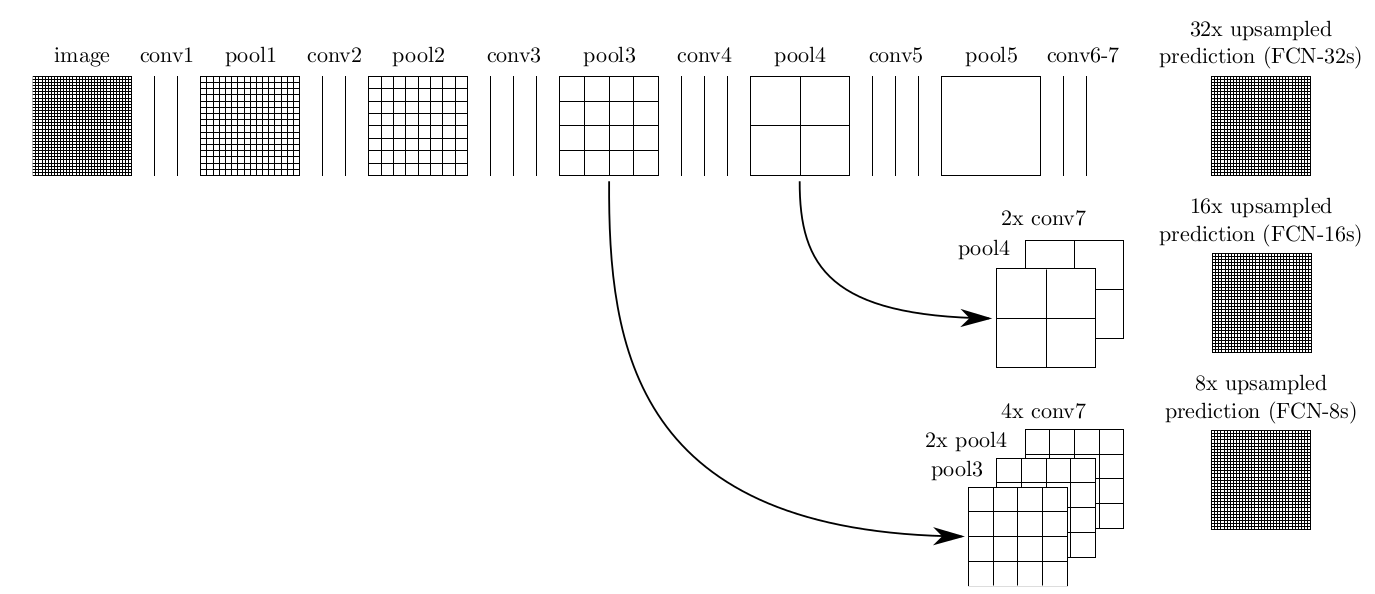}
	\caption{FCN-VGG16 skip architecture representation. Extracted from \cite{shelhamer2017fully}.}
	\label{fig:skipping_layers}
\end{figure}

Shelhamer et al. \cite{shelhamer2017fully} extended their previous paper \cite{long2015fully} about a FCN for pixel-wise prediction which adapts a pre-trained classification network with a skip architecture. 
A classification network, like the VGG16, is then extended using skip layers for pixel-wise prediction. Information of pooling layers (pool3 and 4) is combined with the last convolution layer (conv7), obtaining coarse and fine details to correctly predict local information based on the image structure. Figure \ref{fig:skipping_layers} illustrates the skip architecture using VGG16. As we can see, shallow layers are used in combination with more deeper layers to refine the segmentation (producing the FCN16 and FCN8 models), as using only the final layers would result in coarser segmentations (FCN32 model).

\begin{figure}[th]
	\centering
	\subfloat{%
		\includegraphics[width=0.48\linewidth]{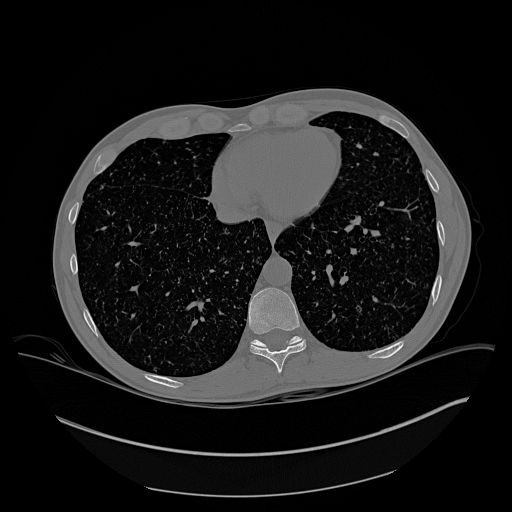}}
	\hfill
	\subfloat{%
		\includegraphics[width=0.48\linewidth]{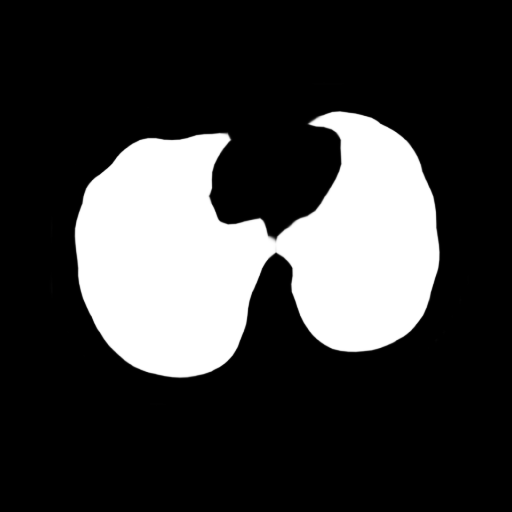}}
	\caption{Example of an input and an output of a FCN. Output image shows the probabilities of being from the lung class rescaled to 0-255.}
	\label{fig:example_output_fcn} 
\end{figure}

The problem is that the output of these layers have different sizes and, since layer fusion is a element-wise operation, both outputs need to have the same shape. They solve this by scaling the smaller output to the larger, and cropping it to keep the same aspect ratio (if padding removed it). FCN8, the model used in this work, has three skip layers, combining information from pool4, pool5 and conv7 layers (shown in Figure \ref{fig:skipping_layers}). After this fusion, the final prediction output is upsampled to the image original resolution and then a per-pixel softmax outputs the probabilities of each pixel being part of the lungs or not. Figure \ref{fig:example_output_fcn} presents a CT slice, used as FCN input, and the corresponding output of the lung class, with its probabilities rescaled to better visualize the result obtained.

Despite the robustness of this technique, an enhancement in this initial segmentation can be employed to refine the outcome. Further improvements of the FCN for semantic segmentation refined their results with the Conditional Random Fields (CRFs), such as the DeepLab \cite{chen2016deeplab}, which also implements the efficient dense CRF proposed in \cite{krahenbuhl2011efficient} by Krahenbuhl and Koltun. They propose a fully connected CRF using Gaussian edge potentials for fast and reliable inference through a mean field approximation based approach, described below.
In the context of CRFs, energy function can be defined as a sum of unary and pairwise potentials:
\begin{equation}
\label{eq:energy_function}
E(x) = \sum_{i}\psi_u(x_i) + \sum_{i,j}\psi_p(x_i, x_j).
\end{equation}

As a post-processing step for FCN segmentation, unary potentials can be treated as the probability maps from softmax's output (more precisely, the negative log of the probability). As for the pairwise potentials, it models relationships between neighborhood, which may be color independent (smoothness kernel) or dependent (appearance kernel). Pairwise potential can be defined as follows:

\begin{equation}
\label{eq:pairwise_potential}
\begin{aligned}
\psi_p(x_i, x_j) = 
&\omega_1 \exp(-\frac{||p_i-p_j||^2}{2\sigma_\alpha^2} -\frac{||I_i-I_j||^2}{2\sigma_\beta^2}) \\ 
& + \omega_2 \exp(-\frac{||p_i-p_j||^2}{2\sigma_\gamma^2})
\end{aligned}
\end{equation}

Then, we find the likely correct assignment (true labels) through a distribution $P(\textbf{x})$, where $P(\textbf{x}) = \exp(-E(\textbf{x}))$. This distribution can be close calculated using a mean-field approximation, which generates a distribution $Q(\textbf{x})$. The idea is to minimize the KL-divergence between $P(\textbf{x})$ and $Q(\textbf{x})$. Each iteration of the mean field executes a message passing step, a compatibility transform and then a local update, until convergence. Through experiments, a optimal value of 10 iterations was proposed. More details about the computation by the mean field approximation are explained in the original article \cite{kohlmann2015automatic}.

Using the probabilities of being in the lung and non-lung classes computed by the FCN, we apply the fully connected CRF to analyze relationships in neighborhood and increase the true positive pixels. The labels assigned by the CRF are used to create the refined segmentation. Details about experiments executed, including optimal parameters found and results achieved, are described in the next section.  

\section{Experimental Results and Discussion}
\label{sec:experiments}

Experiments were realized using two NVIDIA Titan Xp with 12 GB. For HUG-ILD and VESSEL12 datasets, we randomly separated their cases into five folds to assess the reliability of the entire datasets, since a difficult case (i.e. outliers) may be randomly included in the training dataset and not being reported in testing evaluation. Probability of this happening is high in the HUG-ILD dataset, since it has many different pathologies with few cases. Thus, each fold contains approximately 20\% of cases. For each fold, testing subset consists of their cases; as for the training subset, cases of remaining folds were utilized. To correctly parameterize our model, we selected the first fold and applied a 10-fold cross-validation on its training subset, then the best hyper-parameters were applied in the other folds. Training of our model was executed with approximately 19.45 epochs.

Different from \cite{long2015fully}, we employed the Adam optimizer as, through empirical analysis, better results were achieved. Also from initial experiments and based on their work, we determined that the following parameters were optimal for this problem:

\begin{itemize}
	\item Adam epsilon: $1e^{-9}$ (remaining Adam parameters were used as default \cite{kingma2014adam}).
	\item Batch size: The values of 1, 2, 4, 8 and 16 were experimented. Higher values have shown a greatly improvement, converging much faster than a batch size equal to 1, being 8 the optimal choice for both speed and overall results.
	\item Learning rate: Training was faster and better with a learning rate of $1e^{-5}$ than with $1e^{-3}$, $1e^{-4}$ and $1e^{-6}$.
	\item Loss function: We experimented with cross-entropy, Dice and IoU-based losses. Initial experiments shown no substantial improvement, so cross-entropy was selected since its more consolidated.
	\item Weight decay: The default value of $5e^{-4}$ was selected.	
\end{itemize}

Through experiments, we verified that exclusion of the color independent term (smoothness kernel) in the pairwise potential improved segmentation, so we only utilized the appearance kernel in the post-processing step. For the CRF parameters, the values of $\omega_1 = 3$, $\sigma_\alpha=5$ and $\sigma_\beta=26$ contributed most to enhance segmentation. Moreover, inference is executed with 10 iterations (which as also the default value in \cite{krahenbuhl2011efficient}).

Similar to works reported in Section \ref{sec:related_work}, we evaluate our proposed segmentation with the Dice similarity coefficient (DSC) \cite{zou2004statistical}. The Dice coefficient is the spatial overlap score between two sets (e.g. ground truth and proposed segmentation). Given the number of true positives (TP), true negatives (TN), false positives (FP) and false negatives (FN) for each pixel in the CT scan, DSC can be calculated using the Equation \ref{eq:dice} described below:

\begin{equation}
\label{eq:dice}
DSC = \dfrac{2TP}{2TP+FP+FN}
\end{equation}

This coefficient is report for each patient (i.e. 3D score). For each fold, its DSC is reported as the average Dice between all cases. Final score for the dataset is the average Dice of cases from the entire dataset.

Both datasets have their particularities and can influence our model in different ways. Thus, we present a three-part experiment pipeline to evaluate our proposal detailed. These three main experiments are explained in the subsections below, also with their results and discoveries found.

\subsection{Experiment 1: Individual Data Set Training}

In this first experiment, we focus on analyzing the generalization of our model in each dataset individually. The general idea in this experiment is to evaluate each dataset with only their data, for individual study, as one dataset consists mainly of healthy lungs and other of interstitial diseases. Inclusion of data with different characteristics may affect (positively or negatively) overall segmentation. This mixing will be evaluate in latter experiments. 

Table \ref{tab:exp1} shows the scores obtained in the 5-fold evaluation of HUG-ILD and VESSEL12 datasets individually. Mean scores using FCN without and with CRF are reported for each fold, then for the entire dataset. 

\begin{table}[htb]
	\centering
	\caption{Relation of the Dice score obtained for each fold and overall in HUG-ILD and VESSEL12 datasets individually.}
	\label{tab:exp1}
	\begin{tabular}{@{}cllll@{}}
		\toprule
		\multicolumn{1}{l}{} & \multicolumn{2}{l}{HUG-ILD Dataset} & \multicolumn{2}{l}{VESSEL12 Dataset} \\ \midrule
		\# & w/o CRF (\%) & w/ CRF (\%) & w/o CRF (\%) & w/ CRF (\%) \\ \midrule
		1 & $98.57\pm0.89$ & $98.82\pm0.96$ & $98.81\pm0.79$ & $99.11\pm0.60$ \\
		2 & $98.49\pm0.93$ & $98.78\pm0.93$ & $99.18\pm0.28$ & $99.38\pm0.19$ \\
		3 & $98.46\pm0.80$ & $98.73\pm0.77$ & $98.68\pm0.55$ & $98.89\pm0.58$ \\
		4 & $98.06\pm1.12$ & $98.31\pm1.16$ & $99.26\pm0.15$ & $99.45\pm0.14$ \\
		5 & $98.42\pm0.77$ & $98.71\pm0.79$ & $98.94\pm0.58$ & $99.10\pm0.62$ \\ \midrule
		ALL & $98.40\pm0.91$ & $\textbf{98.67}\pm\textbf{0.94}$ & $98.98\pm0.51$ & $\textbf{99.19}\pm\textbf{0.47}$ \\ \bottomrule
	\end{tabular}
\end{table}

\begin{figure}[htb]
	\centering
	\subfloat{%
		\includegraphics[width=0.48\linewidth]{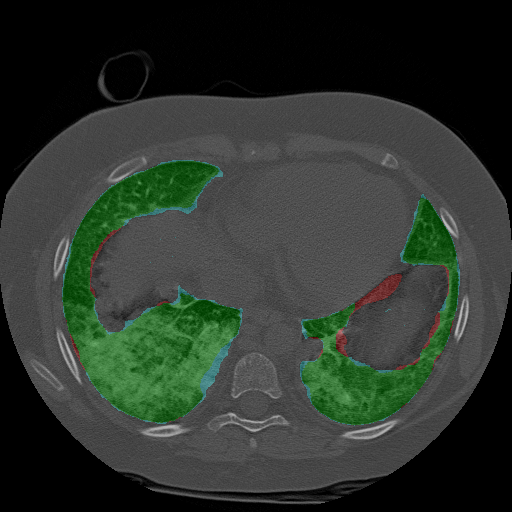}}
	\hfill
	\subfloat{%
		\includegraphics[width=0.48\linewidth]{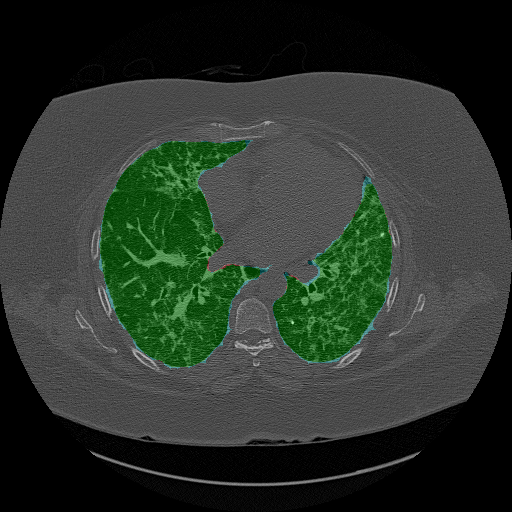}}
	\caption{Resulted segmentation in examples from Figure \ref{fig:example_hug_ild}. Correctly segmented regions are in green, false positives in red and false negatives in cyan.}
	\label{fig:result_example_ild} 
\end{figure}

\begin{figure}[htb]
	\centering
	\subfloat{%
		\includegraphics[width=0.48\linewidth]{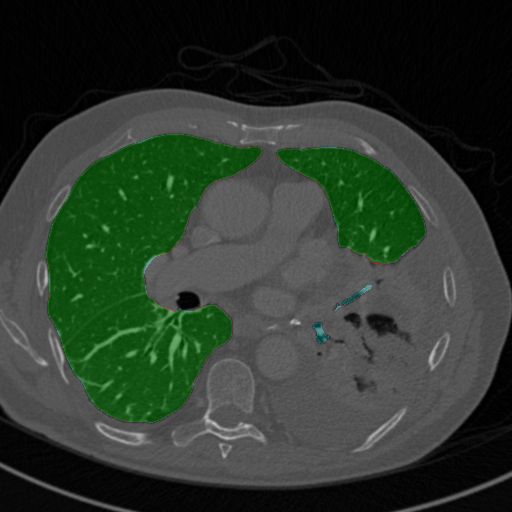}}
	\hfill
	\subfloat{%
		\includegraphics[width=0.48\linewidth]{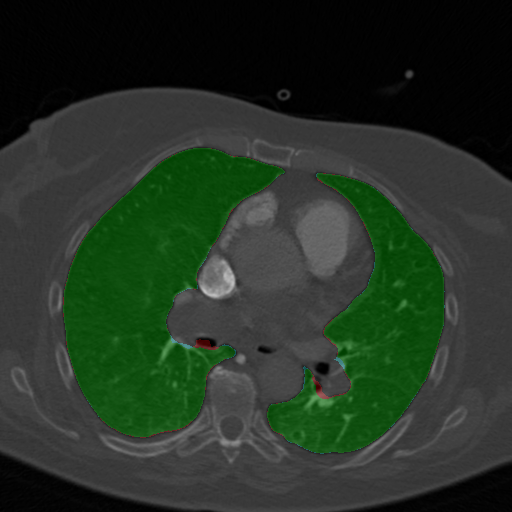}}
	\caption{Segmented slices from the VESSEL12 dataset.}
	\label{fig:result_example_vessel} 
\end{figure}

As we can see, employed of the fully connected CRF improved overall scores for both datasets, with extremely statistically significant ($p < 0.0001$). In the HUG-ILD, Fold \#4 had the lowest mean and highest standard deviation between all folds. Application of CRF reduced the number of cases with $DSC < 97\%$ from five to three, although two remained with scores below $96\%$ ($95.6729\% \rightarrow 95.8967\%$ and $95.6306\% \rightarrow 95.6326\%$). Fold \#1 achieved the best results, with only two cases lower than $97\%$ (but higher than $96\%$). Figure \ref{fig:result_example_ild} illustrates two segmentations from this dataset. Pixels in green are the true positives, red are false positives and pixels in cyan are ground-truth pixels not segmented by our approach. As for the VESSEL12 (examples in Figure \ref{fig:result_example_vessel}), a thing to note is that Fold \#1 not only improved its DSC average with CRF, but also its standard deviation, enhancing greatly the segmentation in its worst case ($97.75\% \rightarrow 98.29\%$).

Comparing with other state-of-the-art works, we improved significantly the results in the HUG-ILD dataset if compared with the P-HNN approach \cite{Harrison2017}, which had a Dice score of $97.9\%\pm1\%$ against $98.40\%\pm0.91\%$ of the FCN only ($p < 0.001$) and $98.67\%\pm0.94\%$ from the FCN+CRF approach ($p < 0.0001$). Despite having a higher mean and lower standard deviation, our FCN+CRF approach in the VESSEL12 dataset was not significantly different from \cite{soliman2017accurate}, which achieved a DSC of $99.0\%\pm0.5\%$ ($p > 0.05$).

\subsection{Experiment 2: Swapping Datasets}

The purpose of this experiment is to analyze variations by using a different testing dataset, not only to check the inter-site differences, but also the peculiarities in a lung pathological dataset. Although the HUG-ILD dataset has slices without pathologies, low quantities may affect results in the VESSEL12 dataset. Table \ref{tab:exp2} shows the results obtained using models from the other dataset with FCN and FCN+CRF.

\begin{table}[ht!b]
	\centering
	\caption{Dice scores swapping the models from HUG-ILD and VESSEL12 datasets.}
	\label{tab:exp2}
	\begin{tabular}{@{}cllll@{}}
		\toprule
		\multicolumn{1}{l}{} & \multicolumn{2}{l}{HUG-ILD Dataset} & \multicolumn{2}{l}{VESSEL12 Dataset} \\ \midrule
		\# & w/o CRF (\%) & w/ CRF (\%) & w/o CRF (\%) & w/ CRF (\%) \\ \midrule
		1 & $96.18\pm5.08$ & $96.48\pm5.09$ & $95.97\pm3.89$ & $96.67\pm3.36$ \\
		2 & $94.85\pm9.85$ & $95.09\pm10.11$ & $97.29\pm0.60$ & $97.82\pm0.62$ \\
		3 & $95.08\pm3.80$ & $95.40\pm3.73$ & $95.38\pm5.39$ & $95.86\pm5.18$ \\
		4 & $95.72\pm5.35$ & $96.00\pm5.39$ & $96.65\pm2.98$ & $97.08\pm2.95$ \\
		5 & $94.93\pm6.90$ & $95.16\pm7.15$ & $96.61\pm3.23$ & $97.02\pm3.17$ \\ \midrule
		ALL & $95.36\pm6.40$ & $\textbf{95.64}\pm\textbf{6.53}$ & $96.39\pm3.30$ & $\textbf{96.89}\pm\textbf{3.14}$ \\ \bottomrule
	\end{tabular}
\end{table}

In general, every model from this experiment had inferior scores than the former, although their average did not decreased greatly. As a negative point in convolutional networks, usage of a model trained with highly different samples from the evaluated cases contributed to the high standard deviations, since the model did not learned to predict cases resembling them. This data-dependence can be a problem, though inclusion of sample from new types (e.g. interstitial diseases, tumors, different lung shapes, etc) may reduce or eliminate this behavior. Thus, we present in next subsection an investigation about this solution proposal. 

\subsection{Experiment 3: Combination of datasets}

In our third and final experiment, we combined the datasets to investigate the influences of a great quantity (from VESSEL12) and variability (from HUG-ILD) of thoracic CT slices. As the former has roughly three times the number of slices than the latter, we selected, for each epoch, only one third of the VESSEL12 dataset for training (the entire dataset is used every three epochs), balancing the influence of both datasets. Table \ref{tab:exp3} presents the results from the combined model evaluated in the HUG-ILD and VESSEL12 datasets.

\begin{table}[ht!b]
	\centering
	\caption{Scores with the combined dataset model applied in both datasets.}
	\label{tab:exp3}
	\begin{tabular}{@{}cllll@{}}
		\toprule
		\multicolumn{1}{l}{} & \multicolumn{2}{l}{HUG-ILD Dataset} & \multicolumn{2}{l}{VESSEL12 Dataset} \\ \midrule
		\# & w/o CRF (\%) & w/ CRF (\%) & w/o CRF (\%) & w/ CRF (\%) \\ \midrule
		1 & $98.58\pm1.00$ & $98.79\pm1.07$ & $98.92\pm0.64$ & $99.16\pm0.54$ \\
		2 & $98.30\pm1.02$ & $98.57\pm1.05$ & $99.02\pm0.44$ & $99.32\pm0.25$ \\
		3 & $98.48\pm0.67$ & $98.75\pm0.66$ & $98.66\pm0.38$ & $98.95\pm0.44$ \\
		4 & $98.23\pm1.10$ & $98.47\pm1.16$ & $99.13\pm0.12$ & $99.36\pm0.13$ \\
		5 & $98.28\pm0.81$ & $98.61\pm0.85$ & $98.95\pm0.41$ & $99.16\pm0.45$ \\ \midrule
		ALL & $98.37\pm0.93$ & $\textbf{98.64}\pm\textbf{0.96}$ & $98.94\pm0.41$ & $\textbf{99.19}\pm\textbf{0.37}$ \\ \bottomrule
	\end{tabular}
\end{table}

\begin{figure*}[htb]
	\centering
	\subfloat[]{%
		\includegraphics[width=0.24\linewidth]{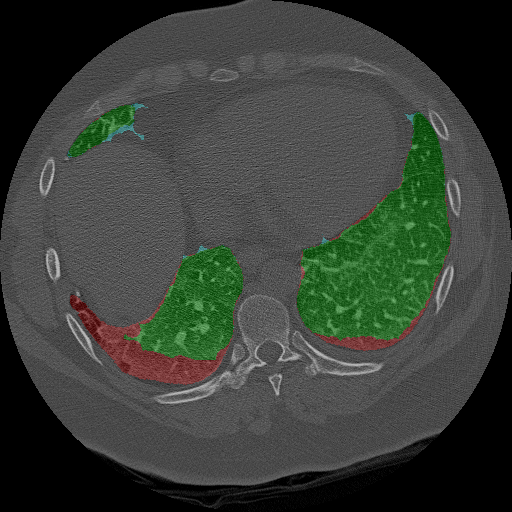}
		\label{fig:worst_results_a}}
	\hfill
	\subfloat[]{%
		\includegraphics[width=0.24\linewidth]{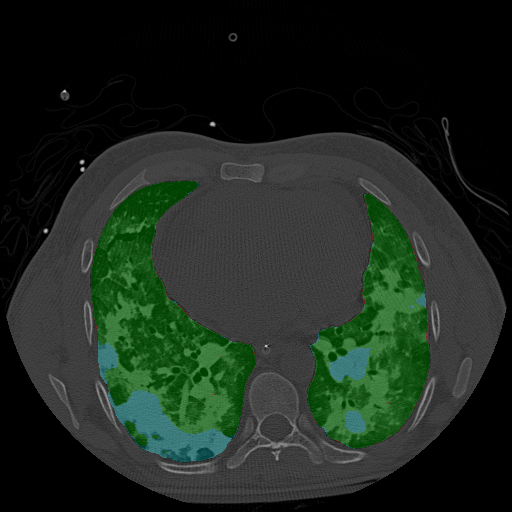}
		\label{fig:worst_results_b}}
	\hfill
	\subfloat[]{%
		\includegraphics[width=0.24\linewidth]{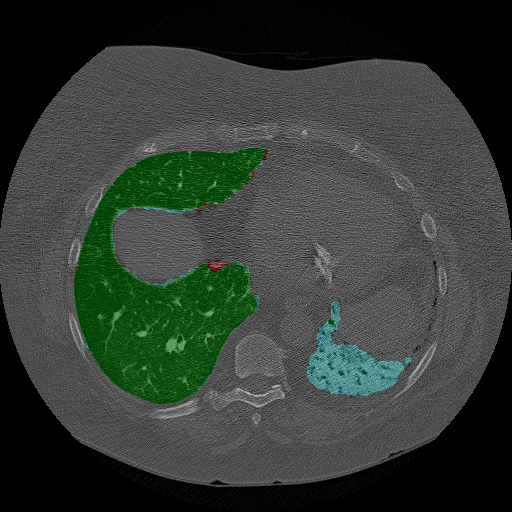}
		\label{fig:worst_results_c}}
	\hfill	
	\subfloat[]{%
		\includegraphics[width=0.24\linewidth]{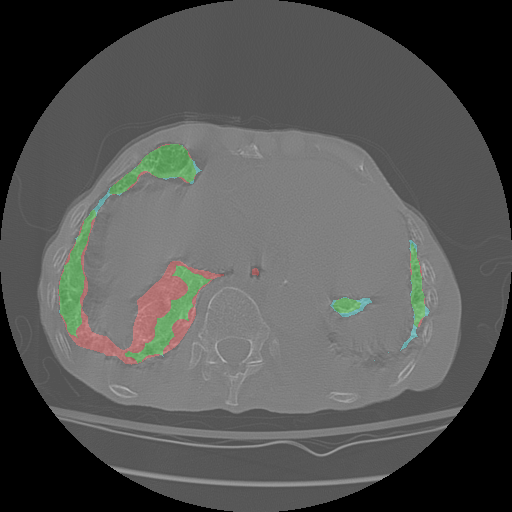}
		\label{fig:worst_results_d}}
	\\
	\subfloat[]{%
		\includegraphics[width=0.24\linewidth]{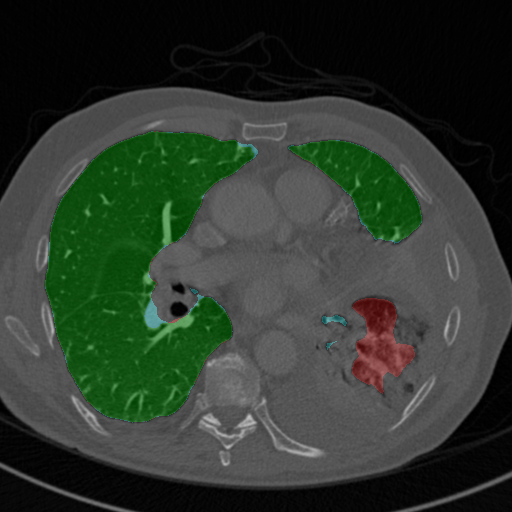}
		\label{fig:worst_results_e}}
	\hfill
	\subfloat[]{%
		\includegraphics[width=0.24\linewidth]{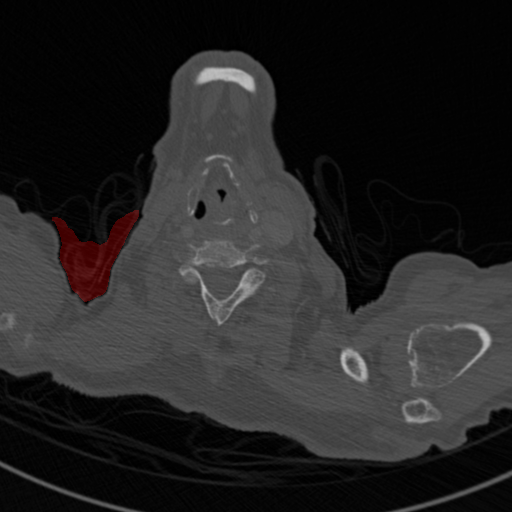}
		\label{fig:worst_results_f}}
	\hfill
	\subfloat[]{%
		\includegraphics[width=0.24\linewidth]{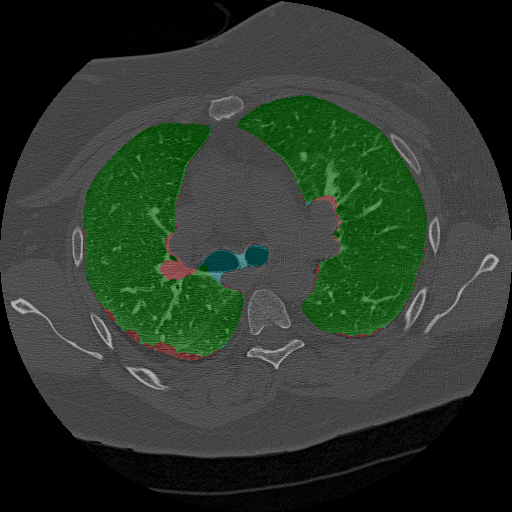}
		\label{fig:worst_results_g}}
	\hfill
	\subfloat[]{%
		\includegraphics[width=0.24\linewidth]{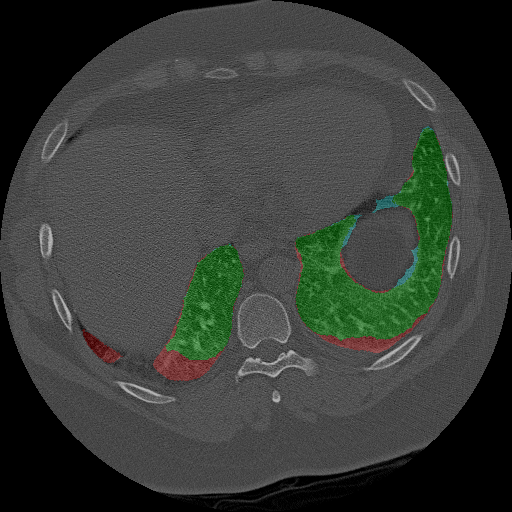}
		\label{fig:worst_results_h}}
	\\
	\subfloat[]{%
		\includegraphics[width=0.24\linewidth]{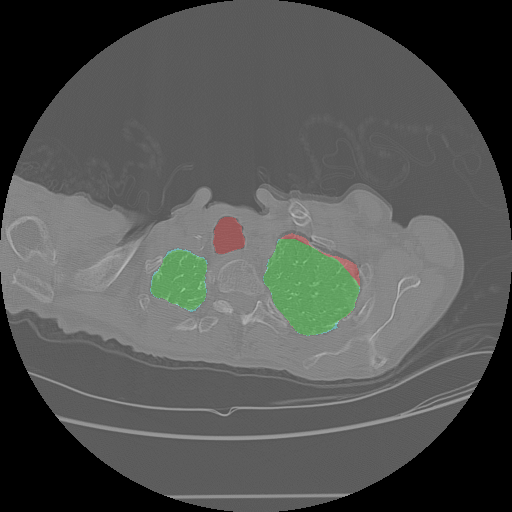}
		\label{fig:worst_results_i}}
	\hfill
	\subfloat[]{%
		\includegraphics[width=0.24\linewidth]{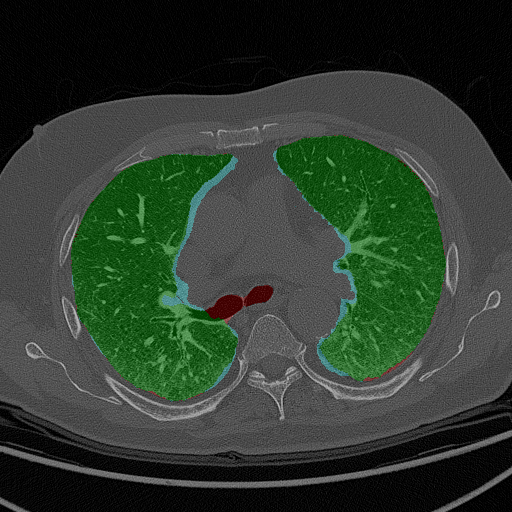}
		\label{fig:worst_results_j}}
	\hfill
	\subfloat[]{%
		\includegraphics[width=0.24\linewidth]{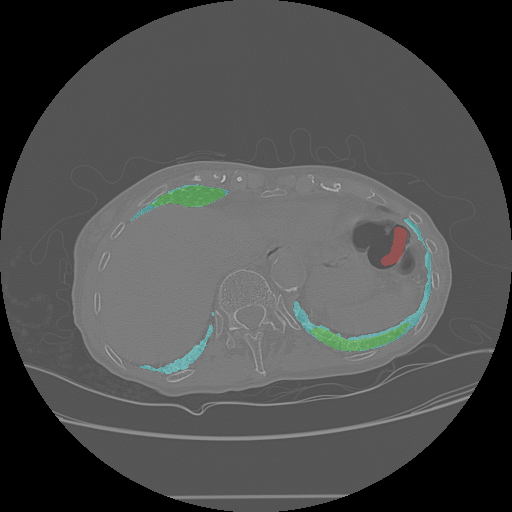}
		\label{fig:worst_results_k}}
	\hfill
	\subfloat[]{%
		\includegraphics[width=0.24\linewidth]{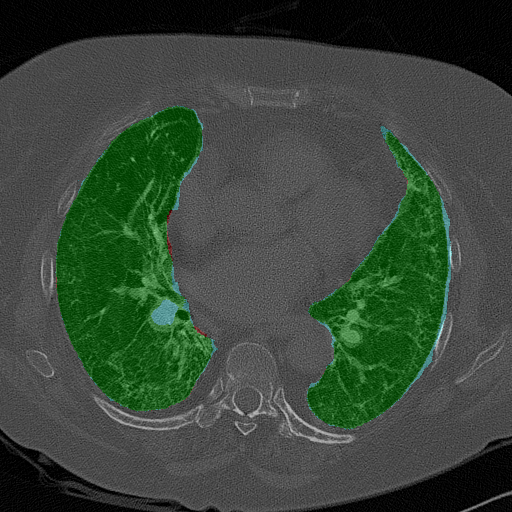}
		\label{fig:worst_results_l}}
	\caption{Some slices which segmentation partially failed, like incorrect segmentation of pathological patterns (\ref{fig:worst_results_b}), inclusion of trachea (\ref{fig:worst_results_e}), borders and thin regions excluded (\ref{fig:worst_results_j} and \ref{fig:worst_results_k}), segmentation of regions outside the thorax (\ref{fig:worst_results_f}).}
	\label{fig:worst_results} 
\end{figure*}

Despite the fact that state-of-the-art results were achieved, some slices were badly segmented. Examples of bad segmentations in our approach are shown in Figure \ref{fig:worst_results}, with correct segmentation in green, and FPs and FNs as red and cyan, respectively. Problems like exclusion of pathological regions, removal of borders and thin regions, and incorporation of incorrect regions, like trachea and parts outside the thorax, were found. Also, uncertain results like regions possibly from the lungs not in the ground-truth (Figures \ref{fig:worst_results_a} and \ref{fig:worst_results_d}) and inclusion of bronchi in one slice and not in another (Figures \ref{fig:worst_results_g} and \ref{fig:worst_results_j}, respectively) may attest as why \cite{Harrison2017} stated errors in HUG-ILD ground-truths.	

Compared with using the generated model from each dataset (first experiment), results using the combined model were similar. Thus, this approach may present more reliable results if used in different datasets, since learned well features from both datasets. Table \ref{tab:final_results_hug_ild} shows results obtained from our experiments compared with the P-HNN approach \cite{Harrison2017}.

\begin{table}[ht!b]
	\centering
	\caption{Comparison of proposed models with state-of-the-art in the HUG-ILD dataset}
	\label{tab:final_results_hug_ild}
	\begin{tabular}{@{}lc@{}}
		\toprule
		Model                       & DSC (in \%)             \\ \midrule
		HUG-ILD FCN                 & $98.40\pm0.91$          \\
		\textbf{HUG-ILD FCN+CRF}    & $\textbf{98.67}\pm\textbf{0.94}$ \\
		VESSEL12 FCN                & $95.36\pm6.40$          \\
		VESSEL12 FCN+CRF            & $95.64\pm6.53$          \\
		ALL FCN                     & $98.37\pm0.93$          \\
		\textbf{ALL FCN+CRF}        & $\textbf{98.64}\pm\textbf{0.96}$ \\
		P-HNN \cite{Harrison2017}   & $97.90\pm1.00$           \\ \bottomrule
	\end{tabular}
\end{table}

Comparison of our models and the work realized by \cite{soliman2017accurate} in the VESSEL12 dataset is presented in Table \ref{tab:final_results_vessel12}.

\begin{table}[ht!b]
	\centering
	\caption{Comparison of proposed models with state-of-the-art in the VESSEL12 dataset}
	\label{tab:final_results_vessel12}
	\begin{tabular}{@{}lc@{}}
		\toprule
		Model                       & DSC (in \%)             \\ \midrule
		HUG-ILD FCN                 & $96.39\pm3.30$          \\
		HUG-ILD FCN+CRF             & $96.89\pm3.14$          \\
		VESSEL12 FCN                & $98.98\pm0.51$          \\
		\textbf{VESSEL12 FCN+CRF}   & $\textbf{99.19}\pm\textbf{0.47}$ \\
		ALL FCN                     & $98.94\pm0.41$          \\
		\textbf{ALL FCN+CRF}        & $\textbf{99.19}\pm\textbf{0.37}$ \\
		Soliman et al. \cite{soliman2017accurate}   & $99.00\pm0.50$           \\ \bottomrule
	\end{tabular}
\end{table}

\section{Conclusion}
\label{sec:conclusion}

A lung segmentation work using fully convolutional networks with post-processing conditional random field applied in ILD and non-ILD datasets is presented in this paper. In this approach, we used the FCN8 with the VGG16 model to generated the lung segmentation, then analyzed the influence of a fully connected CRF with Gaussian edge potentials as a post-processing step. Datasets with different patterns (e.g. healthy lungs, cancerous ones, interstitial diseases) were employed in experiments to evaluate generalization of the proposed models. Experiments demonstrated its capability, obtaining a Dice score of $98.67\%\pm0.94\%$ in the HUG-ILD dataset, outperforming state-of-the-art works ($p < 0.0001$), and obtained similar results with a Dice of $99.19\%\pm0.37\%$ in the VESSEL12 ($p > 0.05$, though had a higher mean and lower standard deviation). Further experiments using different and more variable samples are encouraged to assess its reliability. Also, usage of other classification networks like ResNet and newer versions of the Inception, combined with enhancements in the skip architecture, may lead to improvements.


\section*{Acknowledgment}

This work was supported by CAPES/CNPq. We would like to thank NVIDIA Corporation with the donation of the Titan Xp GPU used in our experiments.



\bibliographystyle{IEEEtran}
\bibliography{IEEEabrv,paper}

\begin{thebibliography}{10}
\providecommand{\url}[1]{#1}
\csname url@samestyle\endcsname
\providecommand{\newblock}{\relax}
\providecommand{\bibinfo}[2]{#2}
\providecommand{\BIBentrySTDinterwordspacing}{\spaceskip=0pt\relax}
\providecommand{\BIBentryALTinterwordstretchfactor}{4}
\providecommand{\BIBentryALTinterwordspacing}{\spaceskip=\fontdimen2\font plus
\BIBentryALTinterwordstretchfactor\fontdimen3\font minus
  \fontdimen4\font\relax}
\providecommand{\BIBforeignlanguage}[2]{{%
\expandafter\ifx\csname l@#1\endcsname\relax
\typeout{** WARNING: IEEEtran.bst: No hyphenation pattern has been}%
\typeout{** loaded for the language `#1'. Using the pattern for}%
\typeout{** the default language instead.}%
\else
\language=\csname l@#1\endcsname
\fi
#2}}
\providecommand{\BIBdecl}{\relax}
\BIBdecl

\bibitem{Torre2015}
\BIBentryALTinterwordspacing
L.~a. Torre, F.~Bray, R.~L. Siegel, J.~Ferlay, J.~Lortet-Tieulent, and
  A.~Jemal, ``{Global cancer statistics, 2012},'' \emph{CA: A Cancer Journal
  for Clinicians}, vol.~65, no.~2, pp. 87--108, mar 2015. [Online]. Available:
  \url{http://doi.wiley.com/10.3322/caac.21262}
\BIBentrySTDinterwordspacing

\bibitem{siegel2018cancer}
\BIBentryALTinterwordspacing
R.~L. Siegel, K.~D. Miller, and A.~Jemal, ``{Cancer statistics, 2018},''
  \emph{CA: A Cancer Journal for Clinicians}, vol.~68, no.~1, pp. 7--30, jan
  2018. [Online]. Available: \url{http://doi.wiley.com/10.3322/caac.21442}
\BIBentrySTDinterwordspacing

\bibitem{el2013computer}
A.~El-Baz, G.~M. Beache, G.~Gimel'farb, K.~Suzuki, K.~Okada, A.~Elnakib,
  A.~Soliman, and B.~Abdollahi, ``Computer-aided diagnosis systems for lung
  cancer: challenges and methodologies,'' \emph{International journal of
  biomedical imaging}, vol. 2013, 2013.

\bibitem{van2013automated}
E.~M. Van~Rikxoort and B.~Van~Ginneken, ``Automated segmentation of pulmonary
  structures in thoracic computed tomography scans: a review,'' \emph{Physics
  in medicine and biology}, vol.~58, no.~17, p. R187, 2013.

\bibitem{weinheimer2011automatic}
O.~Weinheimer, T.~Achenbach, C.~P. Heussel, and C.~D{\"u}ber, ``Automatic lung
  segmentation in {MDCT} images,'' in \emph{Proceedings of the Fourth
  International Workshop on Pulmonary Image Analysis. Toronto}, 2011, pp.
  241--255.

\bibitem{rebouccas2017novel}
P.~P. Rebou{\c{c}}as~Filho, P.~C. Cortez, A.~C. da~Silva~Barros, V.~H.~C.
  Albuquerque, and J.~M.~R. Tavares, ``Novel and powerful {3D} adaptive crisp
  active contour method applied in the segmentation of {CT} lung images,''
  \emph{Medical image analysis}, vol.~35, pp. 503--516, 2017.

\bibitem{sun2012automated}
S.~Sun, C.~Bauer, and R.~Beichel, ``Automated {3-D} segmentation of lungs with
  lung cancer in {CT} data using a novel robust active shape model approach,''
  \emph{IEEE transactions on medical imaging}, vol.~31, no.~2, pp. 449--460,
  2012.

\bibitem{nakagomi2013multi}
K.~Nakagomi, A.~Shimizu, H.~Kobatake, M.~Yakami, K.~Fujimoto, and K.~Togashi,
  ``Multi-shape graph cuts with neighbor prior constraints and its application
  to lung segmentation from a chest {CT} volume,'' \emph{Medical image
  analysis}, vol.~17, no.~1, pp. 62--77, 2013.

\bibitem{dai2015novel}
S.~Dai, K.~Lu, J.~Dong, Y.~Zhang, and Y.~Chen, ``A novel approach of lung
  segmentation on chest {CT} images using graph cuts,'' \emph{Neurocomputing},
  vol. 168, pp. 799--807, 2015.

\bibitem{Harrison2017}
A.~P. Harrison, Z.~Xu, K.~George, L.~Lu, R.~M. Summers, and D.~J. Mollura,
  ``Progressive and multi-path holistically nested neural networks for
  pathological lung segmentation from {CT} images,'' in \emph{Medical Image
  Computing and Computer-Assisted Intervention − MICCAI 2017: 20th
  International Conference, Quebec City, QC, Canada, September 11-13, 2017,
  Proceedings, Part III}.\hskip 1em plus 0.5em minus 0.4em\relax Springer
  International Publishing, 2017, pp. 621--629.

\bibitem{lassen2011lung}
B.~Lassen, J.-M. Kuhnigk, M.~Schmidt, S.~Krass, and H.-O. Peitgen, ``Lung and
  lung lobe segmentation methods at fraunhofer {MEVIS},'' in \emph{Fourth
  international workshop on pulmonary image analysis}, vol. 2011, 2011, pp.
  185--200.

\bibitem{mansoor2014generic}
A.~Mansoor, U.~Bagci, Z.~Xu, B.~Foster, K.~N. Olivier, J.~M. Elinoff, A.~F.
  Suffredini, J.~K. Udupa, and D.~J. Mollura, ``A generic approach to
  pathological lung segmentation,'' \emph{IEEE transactions on medical
  imaging}, vol.~33, no.~12, pp. 2293--2310, 2014.

\bibitem{zhou2014automated}
S.~Zhou, Y.~Cheng, and S.~Tamura, ``Automated lung segmentation and smoothing
  techniques for inclusion of juxtapleural nodules and pulmonary vessels on
  chest {CT} images,'' \emph{Biomedical Signal Processing and Control},
  vol.~13, pp. 62--70, 2014.

\bibitem{shen2015automated}
S.~Shen, A.~A. Bui, J.~Cong, and W.~Hsu, ``An automated lung segmentation
  approach using bidirectional chain codes to improve nodule detection
  accuracy,'' \emph{Computers in biology and medicine}, vol.~57, pp. 139--149,
  2015.

\bibitem{noor2015automatic}
N.~M. Noor, J.~C. Than, O.~M. Rijal, R.~M. Kassim, A.~Yunus, A.~A. Zeki,
  M.~Anzidei, L.~Saba, and J.~S. Suri, ``Automatic lung segmentation using
  control feedback system: morphology and texture paradigm,'' \emph{Journal of
  medical systems}, vol.~39, no.~3, p.~22, 2015.

\bibitem{hosseini20163}
E.~Hosseini-Asl, J.~M. Zurada, G.~Gimel’farb, and A.~El-Baz, ``{3-D} lung
  segmentation by incremental constrained nonnegative matrix factorization,''
  \emph{IEEE Transactions on Biomedical Engineering}, vol.~63, no.~5, pp.
  952--963, 2016.

\bibitem{soliman2017accurate}
A.~Soliman, F.~Khalifa, A.~Elnakib, M.~A. El-Ghar, N.~Dunlap, B.~Wang,
  G.~Gimel’farb, R.~Keynton, and A.~El-Baz, ``Accurate lungs segmentation on
  {CT} chest images by adaptive appearance-guided shape modeling,'' \emph{IEEE
  transactions on medical imaging}, vol.~36, no.~1, pp. 263--276, 2017.

\bibitem{depeursinge2012building}
A.~Depeursinge, A.~Vargas, A.~Platon, A.~Geissbuhler, P.-A. Poletti, and
  H.~M{\"u}ller, ``Building a reference multimedia database for interstitial
  lung diseases,'' \emph{Computerized medical imaging and graphics}, vol.~36,
  no.~3, pp. 227--238, 2012.

\bibitem{rudyanto2014comparing}
R.~D. Rudyanto, S.~Kerkstra, E.~M. Van~Rikxoort, C.~Fetita, P.-Y. Brillet,
  C.~Lefevre, W.~Xue, X.~Zhu, J.~Liang, {\.I}.~{\"O}ks{\"u}z \emph{et~al.},
  ``Comparing algorithms for automated vessel segmentation in computed
  tomography scans of the lung: the {VESSEL12} study,'' \emph{Medical image
  analysis}, vol.~18, no.~7, pp. 1217--1232, 2014.

\bibitem{long2015fully}
J.~Long, E.~Shelhamer, and T.~Darrell, ``Fully convolutional networks for
  semantic segmentation,'' in \emph{Proceedings of the IEEE Conference on
  Computer Vision and Pattern Recognition}, 2015, pp. 3431--3440.

\bibitem{shelhamer2017fully}
E.~Shelhamer, J.~Long, and T.~Darrell, ``Fully convolutional networks for
  semantic segmentation,'' \emph{IEEE transactions on pattern analysis and
  machine intelligence}, vol.~39, no.~4, pp. 640--651, 2017.

\bibitem{chen2016deeplab}
L.-C. Chen, G.~Papandreou, I.~Kokkinos, K.~Murphy, and A.~L. Yuille, ``Deeplab:
  Semantic image segmentation with deep convolutional nets, atrous convolution,
  and fully connected crfs,'' \emph{arXiv preprint arXiv:1606.00915}, 2016.

\bibitem{krahenbuhl2011efficient}
P.~Kr{\"a}henb{\"u}hl and V.~Koltun, ``Efficient inference in fully connected
  {CRFs} with gaussian edge potentials,'' in \emph{Advances in neural
  information processing systems}, 2011, pp. 109--117.

\bibitem{kohlmann2015automatic}
P.~Kohlmann, J.~Strehlow, B.~Jobst, S.~Krass, J.-M. Kuhnigk, A.~Anjorin,
  O.~Sedlaczek, S.~Ley, H.-U. Kauczor, and M.~O. Wielp{\"u}tz, ``Automatic lung
  segmentation method for mri-based lung perfusion studies of patients with
  chronic obstructive pulmonary disease,'' \emph{International journal of
  computer assisted radiology and surgery}, vol.~10, no.~4, pp. 403--417, 2015.

\bibitem{kingma2014adam}
D.~Kingma and J.~Ba, ``Adam: A method for stochastic optimization,''
  \emph{arXiv preprint arXiv:1412.6980}, 2014.

\bibitem{zou2004statistical}
K.~H. Zou, S.~K. Warfield, A.~Bharatha, C.~M. Tempany, M.~R. Kaus, S.~J. Haker,
  W.~M. Wells, F.~A. Jolesz, and R.~Kikinis, ``Statistical validation of image
  segmentation quality based on a spatial overlap index 1: Scientific
  reports,'' \emph{Academic radiology}, vol.~11, no.~2, pp. 178--189, 2004.

\end{thebibliography}
%



\end{document}